\documentclass[conference]{IEEEtran}
\IEEEoverridecommandlockouts
\usepackage{cite}
\usepackage{amsmath,amssymb,amsfonts}
\usepackage{algorithm}
\usepackage{algpseudocode}

\usepackage{graphicx}
\usepackage{textcomp}
\usepackage{xcolor}
\usepackage{soul}
\usepackage{scalerel}
\usepackage{stackengine}
\usepackage{multirow}
\usepackage{float}
\usepackage{booktabs}
\usepackage{multirow}
\usepackage{makecell}

\usepackage{color}

\def\BibTeX{{\rm B\kern-.05em{\sc i\kern-.025em b}\kern-.08em
    T\kern-.1667em\lower.7ex\hbox{E}\kern-.125emX}}

\usepackage[colorlinks=true,linkcolor=red,citecolor=blue, urlcolor=blue]{hyperref}
\usepackage{tabularray}

\newlength{\blob}
\begin{document}

\title{Self Pre-training with Topology- and Spatiality-aware Masked Autoencoders for \\3D Medical Image Segmentation}

\author{%
\IEEEauthorblockN{Pengfei Gu$^{*}$}
\IEEEauthorblockA{\textit{University of Texas Rio Grande Valley}\\
Edinburg, TX 78539, USA\\
pengfei.gu01@utrgv.edu}
\and
\IEEEauthorblockN{Huimin Li$^{*}$\thanks{$*$ Equal contribution}}
\IEEEauthorblockA{\textit{University of Texas Rio Grande Valley}\\
Edinburg, TX 78539, USA\\
huimin.li01@utrgv.edu}
\and
\IEEEauthorblockN{Yejia Zhang}
\IEEEauthorblockA{\textit{University of Notre Dame}\\
Notre Dame, IN 46556, USA\\
chazhang0310@gmail.com}
\and\hspace{3cm}\and
\IEEEauthorblockN{Chaoli Wang}
\IEEEauthorblockA{\textit{University of Notre Dame}\\
Notre Dame, IN 46556, USA\\
chaoli.wang@nd.edu}
\and
\IEEEauthorblockN{Danny Z. Chen}
\IEEEauthorblockA{\textit{University of Notre Dame}\\
Notre Dame, IN 46556, USA\\
dchen@nd.edu}
}%

\maketitle

\begin{abstract}
Masked Autoencoders (MAEs) have been shown to be effective in pre-training Vision Transformers (ViTs) for natural and medical image analysis problems.  
By reconstructing missing pixel/voxel information in visible patches, a ViT encoder can aggregate contextual information for downstream tasks. 
But, existing MAE pre-training methods, which were specifically developed with the ViT architecture, lack the ability to capture geometric shape and spatial information, which is critical for medical image segmentation tasks.
In this paper, we propose a novel extension of known MAEs for self pre-training (i.e., models pre-trained on the same target dataset) for 3D medical image segmentation.
(1) We propose a new topological loss to preserve geometric shape information by computing topological signatures of both the input and reconstructed volumes, learning geometric shape information.
(2) We introduce a pre-text task that predicts the positions of the centers and eight corners of 3D crops, enabling the MAE to aggregate spatial information.
(3) We extend the MAE pre-training strategy to a hybrid state-of-the-art (SOTA) medical image segmentation architecture and co-pretrain it alongside the ViT.
(4) We develop a fine-tuned model for downstream segmentation tasks by complementing the pre-trained ViT encoder with our pre-trained SOTA model.
Extensive experiments on five public 3D segmentation datasets show the effectiveness of our new approach.
\end{abstract}


\section{Introduction}
\label{sec:introduction}
Accurate segmentation of medical images is critical for medical analysis and applications such as diagnosis, treatment planning, and research.
While many deep learning (DL) models (e.g., ~\cite{adame2025topo,zhang2024testfit}) have demonstrated impressive performances in medical image segmentation, 
such methods still face several key challenges. 
One challenge is the scarcity of high-quality labeled medical images for model training, due to high costs and expertise needed for data collection and annotation.
Another challenge is annotation errors, as labeling 3D medical images can be very time-consuming and error-prone. 

\begin{figure}[h!]
    \centering
    \includegraphics[width=0.45\columnwidth]{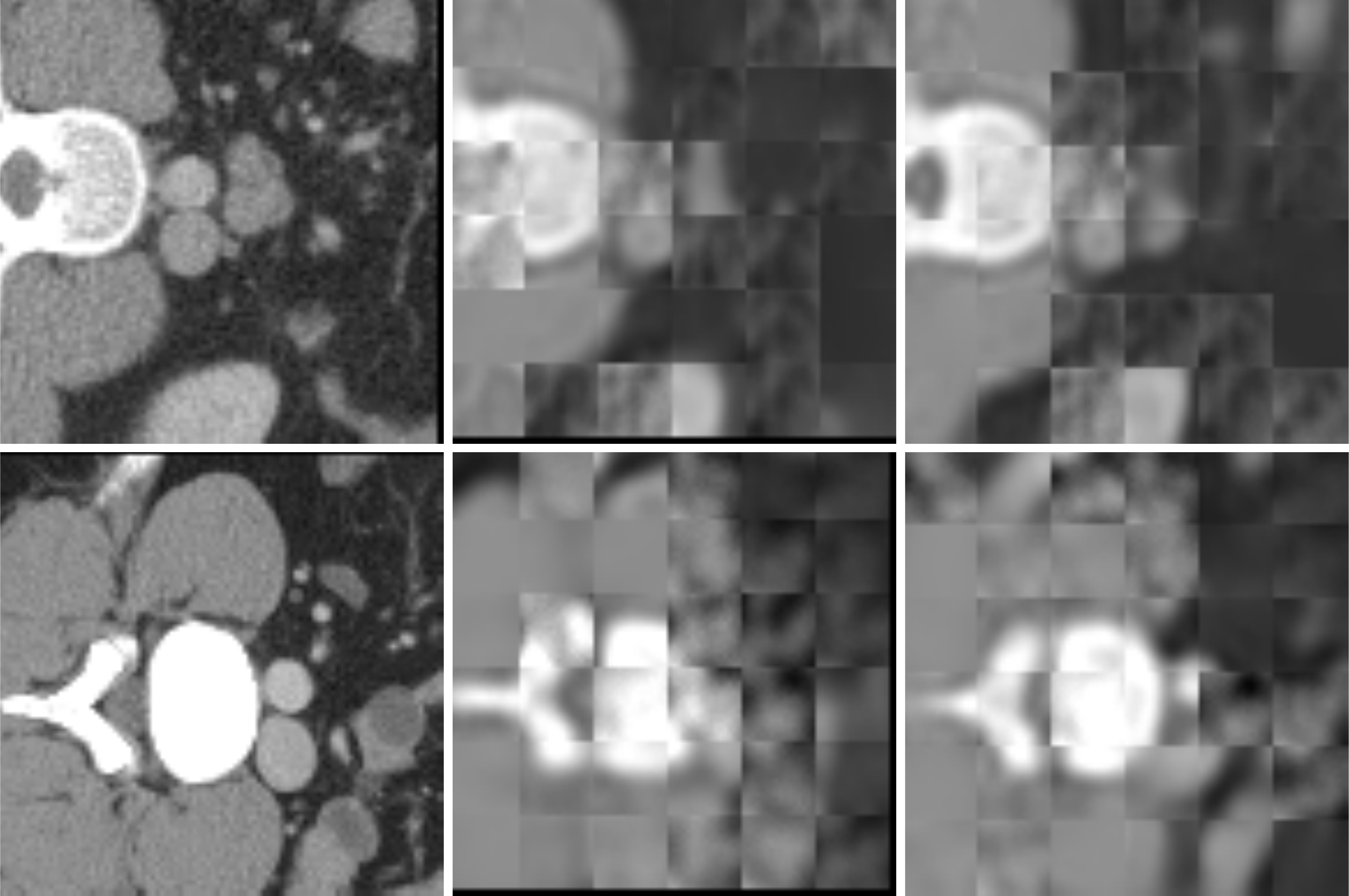}
    \caption{Illustrating the effect of our proposed topological loss. Left: raw image examples of the Synapse CT dataset; middle: reconstructed images with the mean squared error (MSE) loss~\cite{zhou2022self}; right: reconstructed images with a combination of the MSE and proposed topological losses.}
    \label{fig:issues}
\end{figure}

Self-supervised learning (SSL), a technique that leverages pre-text tasks to derive useful visual representations from unlabeled data, offers a promising avenue to combat the challenge of label scarcity.
One representative methodology for SSL is Masked Autoencoders (MAEs)~\cite{he2022masked}. 
Specifically, MAE learns to reconstruct the missing pixels after randomly masking a certain fraction (e.g., 75\%) of patches of the input images. 
In the medical image segmentation area, MAE pre-training has also been found to be effective (e.g., UNETR + MAE~\cite{zhou2022self}).
Although simple and effective, there are still several limitations. 
First, geometric shape information (i.e., contextual information on the overall shapes of objects), which is critical for improving segmentation performance, is not captured well (e.g., see Fig.~\ref{fig:issues}).
Second, global spatial information is not well explored since 
the focus has been on reconstructing information from the masked local sub-volumes, possibly neglecting the global context information of the target objects as a whole.
Third, the MAE pre-training strategy (i.e., learning representations by reconstructing missing patches from masked image input) is not exploited well with various common medical image segmentation architectures, e.g., those based on convolutional neural networks (CNNs) or hybrid models. This is primarily because MAE was developed using the Vision Transformer (ViT)~\cite{dosovitskiy2020image} architecture, potentially restricting its adaptability and effectiveness with other architectures.

To address these limitations, we propose a novel extension of MAEs for self pre-training for 3D medical image segmentation. 
(I) We 
extract geometric shape information by exploiting multi-scale topological features (e.g., connected components, cycles/loops, and voids). 
Our method utilizes cubical complexes~\cite{kaji2020cubical} to compute topological signatures of both the input and reconstructed volumes, and employs an optimal transport distance (the 2-Wasserstein distance) to derive a new topological loss. 
Our topology-aware loss is fully differentiable, computationally efficient, can be added to any neural network, and is applicable to 2D/3D images.
(II) We propose a pre-text task to predict the positions of multiple key points of crops, enabling the model to aggregate spatial information. 
Specifically, our method predicts the positions of nine points (the center and eight corners) of a 3D crop in the input volume.
By learning where the crops are located in the input volume, the model can capture global spatial information.
(III) We extend the MAE pre-training strategy to a hybrid state-of-the-art (SOTA) medical image segmentation architecture, UNETR++~\cite{shaker2022unetr++}, and co-pretrain UNETR++ alongside the ViT.
Specifically, masked crops are processed separately by both ViT and UNETR++ to reconstruct the associated missing patches.
Reconstruction consistency loss and spatial consistency loss (derived from the pre-text task) are employed to connect the two different types of architectures in pre-training, enhancing their representation learning capability.

Following~\cite{zhou2022self}, our method is performed on self pre-training paradigms (i.e., models pre-trained on the same target dataset).
In the self pre-training stage, we randomly mask a fraction (e.g., 50\%) of patches of the image crops. 
The masked crops are then processed independently by a ViT model and a UNETR++ model, which are pre-trained with our proposed topological loss, pre-text task that predicts the positions of 9 key points of crops, and spatial and reconstruction consistency losses, learning the geometric shape and global spatial information and enhancing the representation learning capability. 
In the fine-tuning stage, the pre-trained ViT encoder is complemented with the pre-trained UNETR++ model, which is then fine-tuned for the target segmentation task. 
A fusion module is utilized to fuse the scale-wise features from both the pre-trained ViT encoder and UNETR++ encoder.

Our main contributions are summarized as follows: 
(1) We propose a new topological loss and introduce a pre-text task for MAEs to learn geometric shape and spatial information.
(2) We extend the MAE pre-training strategy to a hybrid SOTA medical image segmentation architecture and co-pretrain it alongside ViT.
(3) We develop a fine-tuned model for downstream segmentation tasks, and demonstrate the effectiveness of our new approach on five public 3D segmentation datasets.

\begin{figure}[h!]
    \centering
    \includegraphics[width=0.88\linewidth]{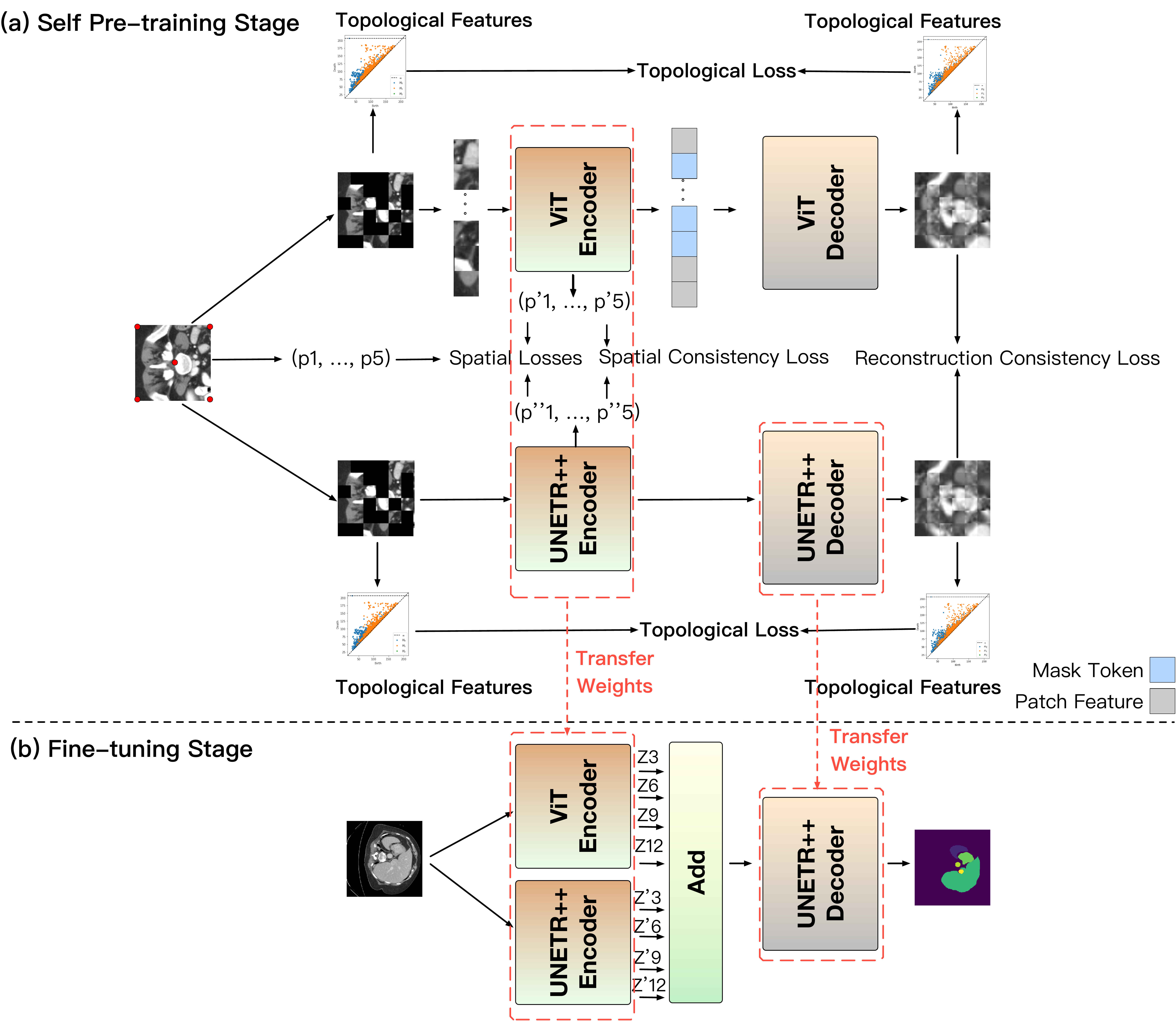}
    \caption{An overview of our proposed pipeline.}
    \label{fig:overview}
\end{figure}

\section{Method}\label{sec:method}
Fig.~\ref{fig:overview} presents an overview of our proposed pipeline, 
which contains four main components: (1) a topological loss that aims at implicitly extracting geometric shape information by exploiting multi-scale topological features;
(2) a pre-text task that captures global spatial information by predicting the positions of 9 key points of 3D crops in the input volume;
(3) a spatial consistency loss and a reconstruction consistency loss that enhance the representation learning capability of both the ViT and UNETR++ models by aligning the reconstructed images at both spatial and image levels;
(4) a fine-tuned model for improving the downstream segmentation performance.


\subsection{Capturing the Topology of Input Volumes}\label{subsec:topo}
Given a 3D image $I$, we represent $I$ with a \textit{cubical complex} $C$.
Typically, the cubical complex $C$ takes each voxel of $I$ as an individual vertex and contains connectivity information on vertex neighbourhoods via edges, squares, and their higher-dimensional counterparts~\cite{kaji2020cubical,gu2025topoimages}.
In this work, we use \textit{persistent homology} (PH)~\cite{edelsbrunner2002topological} to extract topological features of different dimensions from 
$C$, including connected components ($0$-D), cycles/loops ($1$-D), and voids ($2$-D). 
PH combines the homology of super-level sets by sweeping a threshold function through the entire real numbers.
Specifically, for a threshold value $\tau \in \mathbb{R}$, a cubical complex is defined as:
$C^{(\tau)} := \{x\in I \ | \ f(x) \geq \tau\},
$
where $f(x)$ is the voxel value of $x$. 
When sweeping the threshold, the topology changes only at a finite number of values, $\tau_1\geq \tau_2 \geq \cdots \geq \tau_{m-1} \geq \tau_m$, and we obtain a sequence of nested cubical complexes, $\emptyset  \subseteq C^{(\tau_1)} \subseteq C^{(\tau_2)} \subseteq \cdots \subseteq C^{(\tau_{m-1})}\subseteq C^{(\tau_m)} = I$, which forms the \textit{super-level set filtration}.
PH tracks topological features across all the complexes in this filtration, representing each feature as a tuple $(\tau_i, \tau_j)$ with $\tau_i \geq \tau_j,$ indicating the cubical complex in which a feature appears and disappears, respectively. 
For example, a $0$-D tuple $(\tau_i, \tau_j)$ represents a connected component that appears at threshold $\tau_i$ and disappears at threshold $\tau_j$.
The tuples of the $k$-D ($0\leq k\leq 2$) features are saved in the \textit{$k$-th persistence diagram} $D_I^{k}$, which is a multi-scale shape descriptor of all topological features of the 3D image $I$.

\textbf{Comparing Persistence Diagrams.}
Given two persistence diagrams $D$ and $D'$, we use the \textit{2-Wasserstein distance}
as a metric to measure their similarity or distance, defined as: 
$
W_2(D,D') := \left(\inf_{\eta: D \rightarrow D'}\sum_{x \in D}||x- \eta(x)||_{\infty}^{2}\right)^{\frac{1}{2}},
$
where $\eta(\cdot)$ denotes a bijection. Note that this equation can be solved by using an optimal transport algorithm, and we use cubical Ripser~\cite{kaji2020cubical} to compute PHs from volumes.

\textbf{Constructing Topological Loss.}
Given an input volume $I$ and a reconstructed volume $I'$, our new topology-aware loss is defined as:
$
    \mathcal{L}_{topo}(I, I') = \left(\sum_{i=0}^{2}( W_2(D_{I}^{i}, D_{I'}^{i}))^2\right)^{\frac{1}{2}},
$
where $W_2(\cdot,\cdot)$ denotes the 2-Wasserstein distance, and $D_{I}^{i}$ and $D_{I'}^{i}$ are the $i$-D persistence diagrams of 
$I$ and 
$I'$, respectively. 
Note that our proposed topological loss differs from that in~\cite{waibel2022capturing}. First, we extract topological features from 3D images, not from the segmentation. Second, we use the topological loss for self pre-training 
and medical image segmentation, not for 3D reconstruction tasks.

\subsection{Exploiting Global Spatial Information}\label{subsec:pre-text}
MAE~\cite{he2022masked} and UNETR + MAE~\cite{zhou2022self} lack the ability to learn global spatial information that is vital to 3D medical image segmentation for two reasons: (1) The positional embedding encodes only local position information for each patch, and 
(2) the methods focus only on low-level patch matching with a local mean squared error (MSE) loss. 
To address these limitations, we propose a novel pre-text task that complements the known methods with global spatial information.
Specifically, the pre-text task aims to predict the positions of 9 key points (the center and eight corners) of 3D crops in the input volume. 
We attain this by adding two prediction heads to the ViT and UNETR++ encoders.
The two prediction heads share the same architecture that consists of a convolutional layer, a two-layer multilayer perceptron (MLP) with 256 hidden dimensions, and a $\tanh$ activation function. 
This design enables the ViT and UNETR++ encoders to learn global spatial representations.

\textbf{Constructing Spatial Loss.}
We denote the 9 key points of a 3D crop as $(p_1, p_2$, $\ldots, p_{9})$, where $p_9$ is for the crop center, and each $p_i = (x_i, y_i, z_i)$.
Given the ground truth (GT) and the prediction of the 9 positions, $P= (p_1, p_2, \ldots, p_{9})$ and $P'=(p'_1, p'_2, \ldots, p'_{9})$, the spatial loss is defined as:
$
    \mathcal{L}_{spa}(P, P') =\mathcal{L}_{MSE}(P, P'),
$
where $\mathcal{L}_{MSE}$ is the MSE loss.
Our spatial loss definition and implementation are different from those in~\cite{zhang2022point}.

\subsection{Co-pretraining the ViT and UNETR++ Models}\label{subsec:consistency}
As illustrated in Fig.~\ref{fig:overview}, the masked crops are processed independently by both the ViT and UNETR++ models.
To co-pretrain both the ViT and UNETR++ models to enhance their representation learning capability, we propose to align the reconstructed images in the spatial and image levels.

\textbf{Constructing Spatial Consistency Loss.}\label{subsec:consis}
Given predictions of 9 key point positions from the ViT and UNETR++ encoders, $P'=(p'_1, p'_2, \ldots, p'_{9})$ and $P''=(p''_1, p''_2, \ldots, p''_{9})$, the spatial consistency loss is defined as:
$
    \mathcal{L}_{spa-consis}(P', P'') =\mathcal{L}_{MSE}(P', P'').
$
The spatial consistency loss aligns the reconstructed images at the spatial level, enhancing the feature learning capability of both the ViT and UNETR++ models.

\textbf{Constructing Reconstruction Consistency Loss.}
Given reconstructed volumes $I'$ and $I''$ by the ViT and UNETR++ models, our reconstruction consistency loss function
computes the MSE between the reconstructed volumes $I'$ and $I''$, as:
$
    \mathcal{L}_{rec-consis}(I', I'') =\mathcal{L}_{MSE}(I',I'').
$
We compute this loss only on masked patches, similar to MAE in~\cite{he2022masked}.
The reconstruction consistency loss aligns the reconstructed images at the image level, further enhancing the representation learning capability of both ViT and UNETR++.

\textbf{The Overall Loss of Self Pre-training.}
The overall loss for a volume crop is:
\begin{multline*}
 \mathcal{L} = (1-\lambda_1)(1-2\lambda_2)\mathcal{L}_{MSE-ViT} + (1-\lambda_1)\lambda_2\mathcal{L}_{topo-ViT}\\
 + (1-\lambda_1)\lambda_2 \mathcal{L}_{spa-Vit} + \lambda_1(1-2\lambda_2)\mathcal{L}_{MSE-UNETR++}\\
 +\lambda_1\lambda_2\mathcal{L}_{topo-UNETR++} +  \lambda_1\lambda_2\mathcal{L}_{spa-UNETR++}\\
 + \lambda_3 \mathcal{L}_{spa-consis} + \lambda_3 \mathcal{L}_{rec-consis},
\end{multline*}
where $\mathcal{L}_{MSE-X}$, $\mathcal{L}_{topo-X}$, and $\mathcal{L}_{spa-X}$ are the reconstruction, topological, and spatial losses, respectively, for the $X$ (either ViT or UNETR++) model, and $\lambda_i$ is a balancing weight.

\subsection{Constructing the Fine-tuned Architecture}\label{subsec:model}
In~\cite{zhou2022self}, the pre-trained ViT encoder weights were transferred to initialize the segmentation encoder, i.e., the UNETR~\cite{hatamizadeh2022unetr} encoder, achieving impressive performance.
Following~\cite{zhou2022self}, we utilize the pre-trained ViT encoder weights and propose to complement the pre-trained ViT encoder with the pre-trained  UNETR++~\cite{shaker2022unetr++} to enhance the performance of downstream segmentation tasks.

As shown in Fig.~\ref{fig:overview}, our fine-tuned model consists of four key components: two pre-trained encoders (the pre-trained ViT and UNETR++ encoders), an add fusion module, and a pre-trained UNETR++ decoder. 
Specifically, the two encoders are employed to capture complementary features, since the ViT encoder is a Transformer-based architecture and the UNETR++ encoder is a convolution-based architecture. 
Then a scale-wise fusion module, which is addition, is used to fuse the scale-wise features from the two different types of encoders.
Finally, a pre-trained UNETR++ decoder is appended to generate the final segmentation.
Our fine-tuned model is called {\bf MAE + UNETR++}, which can effectively leverage the pre-trained ViT encoder to capture high-level semantic information and the pre-trained UNETR++ to better capture fine details and edges, resulting in improved segmentation performance.

\begin{table*}[!t]
   \caption{Segmentation results on the Synapse CT dataset. The best results are marked in {\bf bold}, and the second-best results are \underline{underlined}.}\label{tab:Synapse}
   \vspace{-4mm}
    \centering
    \vspace{0.2cm}
     \scalebox{0.75}{
   \begin{tabular}{l|cc|cccccccc|cc}
   \hline
   \multirow{2}{*}{Method} & \multirow{2}{*}{Params.} & \multirow{2}{*}{FLOPs} & \multirow{2}{*}{Spl} & \multirow{2}{*}{RKid} & \multirow{2}{*}{LKid} & \multirow{2}{*}{Gal} & \multirow{2}{*}{Liv} & \multirow{2}{*}{Sto} & \multirow{2}{*}{Aor} & \multirow{2}{*}{Pan} &\multicolumn{2}{c}{Average}\\
   \cline{12-13}
   &&&&&&&&&&& Dice ($\uparrow$) & HD95 ($\downarrow$) \\
   \hline\hline
   \multirow{1}{*}{U-Net~\cite{ronneberger2015u}}    
                     &---   & ---   & 86.67 &68.60 &77.77 &69.72 &93.43 &75.58 &89.07 &53.98 & 76.85&--- \\
    \multirow{1}{*}{TransUNet~\cite{chen2021transunet}}   
                     &96.07M & 88.91   & 85.08 &77.02 &81.87 &63.16 &94.08 &75.62 &87.23 &55.86  & 77.49&31.69        \\
     \multirow{1}{*}{UNETR~\cite{hatamizadeh2022unetr}}   
                     &92.49M   & 75.76  & 85.00 &84.52 &85.60 &56.30 &94.57 &70.46 &89.80 &60.47   & 78.35&18.59      \\
            \multirow{1}{*}{Swin-UNet~\cite{cao2023swin}}   
                     &---   & ---   & 90.66 &79.61 &83.28 &66.53 &94.29 &76.60 &85.47 &56.58   & 79.13&21.55      \\
            \multirow{1}{*}{MISSFormer~\cite{huang2021missformer}}
                     &---   & ---  & 91.92 &82.00 &85.21 &68.65 &94.41 &80.81 &86.99 &65.67  & 81.96&18.20       \\
            \multirow{1}{*}{Swin UNETR~\cite{hatamizadeh2022swin}}
                     &62.83M   & 384.2   & 95.37 &86.26 &86.99 &66.54 &95.72 &77.01 &91.12 &68.80   & 83.48&10.55      \\
              \multirow{1}{*}{UNETR + MAE~\cite{zhou2022self}}
                     &---   & --- & 90.56 &84.00 &86.37 &\underline{75.25} &95.95 &80.89 &88.92 &65.02 & 83.52&10.24         \\
            \multirow{1}{*}{nnFormer~\cite{zhou2021nnformer}}
                     &150.5M   &213.4  & 90.51 &86.25 &86.57 &70.17 &\underline{96.84} &\underline{86.83} &92.04 &\textbf{83.35}       & 86.57&10.63  \\
            \multirow{1}{*}{UNETR++~\cite{shaker2022unetr++}}
                     &42.96M  & 47.98  & \textbf{95.77} &\underline{87.18} &\underline{87.54} &71.25 &96.42 &86.01 &\underline{92.52} &81.10  & \underline{87.22}&\underline{7.53}\\
   \hline \hline
   \multirow{1}{*}{ MAE + UNETR++ (ours)}
                     &85.96M  &82.49   &\underline{95.68}   &\textbf{89.30} &\textbf{87.64} &\textbf{79.60} &\textbf{96.98} &\textbf{88.47} & \textbf{92.58}& \underline{81.27} & \textbf{88.94 $(1.72\%\uparrow)$} &\textbf{5.89 $(1.64 \uparrow)$} \\\hline 
   \multirow{1}{*} {$p$-values}
                     & \multicolumn{12}{c}{$< 5e-2$ (Dice), $< 1e-2$ (HD95)}   \\\hline 
   \end{tabular}
   }
\end{table*}

\section{Experiments and analysis}
\label{sec:exp}

\subsection{Datasets and Experimental Setup}\label{datasets}
We conduct experiments on five segmentation datasets: Synapse
multi-organ CT segmentation (Synapse CT Dataset)~\cite{landman2015miccai}, 
BTCV multi-organ CT segmentation (BTCV CT Dataset)~\cite{landman2015miccai},
ACDC automated cardiac diagnosis (ACDC)~\cite{bernard2018deep}, and Medical Segmentation Decathlon (MSD) datasets~\cite{antonelli2022medical} for two different segmentation tasks, spleen segmentation and lung segmentation. 
For each experiment, we perform 5 runs using different random seeds and report the average results.
Additionally, we compute $p$-values to ascertain the statistical significance of the results.

{\bf Synapse CT Dataset:} 
This dataset~\cite{landman2015miccai} contains $30$ abdominal CT volumes with $8$ organs. 
Following~\cite{chen2021transunet,zhang2022keep}, we split the dataset randomly into $18$ volumes and $12$ volumes for training and testing, and report the average Dice and $95\%$ Hausdorff distance (HD95) on $8$ abdominal organs: spleen (Spl), right kidney (RKid), left kidney (LKid), gallbladder (Gal), liver (Liv), stomach (Sto), aorta (Aor), and pancreas (Pan).

{\bf BTCV CT Dataset:}: 
This dataset~\cite{landman2015miccai} consists of $30$ abdominal CT volumes with $13$ organs, including 8 organs of the Synapse CT dataset, along with esophagus (Eso), inferior vena cava (IVC), portal and splenic veins (PSV), right adrenal gland (RAG), and left adrenal gland (LAG). 
%
%

%
{\bf ACDC Dataset:} 
This dataset~\cite{bernard2018deep} contains 100 samples, and aims to segment the cavity of the right ventricle, the myocardium of the left ventricle, and the cavity of the left ventricle. Each sample’s labels involve left ventricle (LV), right ventricle (RV), and myocardium (MYO). 
Following~\cite{zhou2021nnformer}, we split the dataset into 70 training samples, 10 validation samples, and 20 test samples.

{\bf MSD Spleen Dataset:} 
This dataset~\cite{antonelli2022medical} contains $41$ CT volumes for spleen segmentation. 
Following~\cite{hatamizadeh2022unetr}, we split the dataset into training, validation, and test sets (80:15:5).
{\bf MSD Lung Dataset:}:  This dataset~\cite{antonelli2022medical} comprises 64 CT volumes for lung cancer segmentation. 
We split the dataset with a 80:20 ratio for training and testing following~\cite{shaker2022unetr++}.

\subsection{Implementation Details}\label{imp}
Our experiments are implemented with PyTorch and MONAI.
The model training is performed on an NVIDIA Tesla V100 Graphics Card with 32GB GPU memory using the AdamW optimizer with a weight decay = $0.005$. 

For the Synapse CT and BTCV CT datasets, we clip the raw values between $-175$ and $250$, normalize the values into the range of [0, 1], and re-sample the spacing to $[1.5, 1.5, 2.0]$. All the models are trained with input images of size $96 \times 96 \times 96$. 
For the ACDC dataset, we re-sample the spacing to $[1.52, 1.52, 6.35]$. All the models are trained with input of size $160 \times 160 \times 16$. 
For the MSD spleen dataset, we clip the raw values between $-57$ and $164$, normalize the values into the range of [0, 1], and re-sample the spacing to $[1.5, 1.5, 2.0]$. All the models are trained with input of size $96 \times 96 \times 96$. 
For the MSD lung dataset, we clip the raw values between $-1000$ and $3071$, normalize the values into the range of [0, 1], and re-sample the spacing to $[1.0, 1.0, 1.0]$. All the models are trained with input of size $192 \times 192 \times 32$.  

We use a learning rate of $6.4e-3$ for self pre-training on all the datasets.
We pre-train on the Synapse CT, BTCV CT, and MSD spleen and lung segmentation datasets with $10,000$ epochs, and on the ACDC dataset with $2,000$ epochs.

For all the downstream segmentation tasks, we use a learning rate of $1e-1$, and fine-tune with $5,000$ epochs for the Synapse CT, BTCV CT, and MSD spleen and lung segmentation datasets, and $1,000$ epochs for the ACDC dataset.
The batch size for each case is set as the maximum size allowed by the GPU. 
We set $\lambda_1=0.5$, $\lambda_2=0.1$, and $\lambda_3=0.1$.
    
\begin{table*}[!t]
   \caption{Segmentation results on the BTCV CT dataset.}\label{tab:btcv}
   \vspace{-4mm}
    \centering
    \vspace{0.2cm}
    \scalebox{0.75}{
   \begin{tabular}{l|ccccccccccccc|cc}
   \hline
   \multirow{1}{*}{Method} & \multirow{1}{*}{Spl} & \multirow{1}{*}{RKid} & \multirow{1}{*}{LKid} & \multirow{1}{*}{Gal} & \multirow{1}{*}{Eso}& \multirow{1}{*}{Liv} & \multirow{1}{*}{Sto} & \multirow{1}{*}{Aor} & \multirow{1}{*}{IVC}& \multirow{1}{*}{PSV}& \multirow{1}{*}{Pan}
   & \multirow{1}{*}{RVG}
   & \multirow{1}{*}{LAG}&\multicolumn{1}{c}{Average Dice ($\uparrow$)}\\
   \hline\hline
    \multirow{1}{*}{UNETR~\cite{hatamizadeh2022unetr}}
                      &90.48 &82.51 & 86.05 &58.23 &71.21 & 94.64 &72.06 &86.57 &76.51 &70.37 & 66.06 &66.25 &63.04  &76.00      \\
     {Swin UNETR~\cite{hatamizadeh2022swin}} 
                      &94.59 &88.97 &92.39 &65.37 &75.43 &95.61 &75.57 &88.28 &81.61 &76.30 &74.52 &68.23 &66.02  &80.44    \\
    \multirow{1}{*}{TransBTS~\cite{wang2021transbts}} 
                      &94.55 &\underline{89.20} & 90.97 &68.38 &75.61 &96.44 & 83.52 & 88.55 & 82.48 & 74.21 &76.02 &67.23 &67.03  &81.31    \\
        \multirow{1}{*}{nnFormer~\cite{zhou2021nnformer}}
                      &94.58 &88.62 &\textbf{93.68} &65.29 &76.22 &96.17 &83.59 &89.09 &80.80 &75.97 &77.87 &70.20 &66.05  &81.62   \\
   \multirow{1}{*}{nnU-Net~\cite{isensee2021nnu}}
              &\textbf{95.95} &88.35 & 93.02 &70.13 & 76.72 &96.51 & \underline{86.79} &88.93 & 82.89 &\textbf{78.51} &\underline{79.60} & \textbf{73.26} &\underline{68.35}  &83.16    \\
            \multirow{1}{*}{UNETR++~\cite{shaker2022unetr++}}
                     &{94.94} &\textbf{91.90} &\underline{93.62} &\underline{70.75} &\underline{77.18} &\underline{95.95} &85.15 &\underline{89.28} &\underline{83.14} &76.91 &77.42 &\underline{72.56} &68.17  &\underline{83.28} \\
   \hline \hline
   \multirow{1}{*}{ MAE + UNETR++ (ours)}
                    &\underline{94.97} &87.93 &87.37 &\textbf{78.46} &\textbf{78.97} &\textbf{96.99} &\textbf{88.31} &\textbf{92.51} &\textbf{89.01} &\underline{76.94} &\textbf{80.18} &69.88 &\textbf{71.48}  & \textbf{84.08 $(0.8\% \uparrow)$}  \\\hline 
   \multirow{1}{*} {$p$-value}
                     & \multicolumn{14}{c}{$< 1e-2$ (Dice)}   \\\hline 
   \end{tabular}
   }
\end{table*}

\begin{table}[!t]
   \caption{Segmentation results on the ACDC dataset. }\label{tab:acdc}
   \vspace{-4mm}
    \centering
    \vspace{0.2cm}
    \scalebox{0.7}{
   \begin{tabular}{l|ccc|cc}
   \hline
   \multirow{1}{*}{Method}  & \multirow{1}{*}{RV} & \multirow{1}{*}{Myo} & \multirow{1}{*}{LV}   &\multicolumn{1}{c}{Average}\\
   \hline\hline
   \multirow{1}{*}{VIT-CUP~\cite{dosovitskiy2020image}}      
                       &81.46 & 70.71 & 92.18 &81.45 \\
   \multirow{1}{*}{R50-VIT-CUP~\cite{dosovitskiy2020image}}      
   &86.07 &  81.88 & 94.75 &87.57 \\
    \multirow{1}{*}{MISSFormer~\cite{huang2021missformer}}
    &86.36 &  85.75 &  91.59 &87.90 \\
    \multirow{1}{*}{UNETR~\cite{hatamizadeh2022unetr}}      
   &85.29 & 86.52 &  94.02 &88.61 \\
   \multirow{1}{*}{TransUNet~\cite{chen2021transunet}}      
                       &88.86 & 84.54 & 95.73 &89.71 \\
   \multirow{1}{*}{Swin-UNet~\cite{cao2023swin}}       
   &88.55 & 85.62 & 95.83 &90.00 \\
   \multirow{1}{*}{LeViT-UNet-384s~\cite{xu2021levit}}      
   &89.55 & 87.64 & 93.76 &90.32 \\
    \multirow{1}{*}{UNETR + MAE~\cite{zhou2022self}}
            &--- (88.44)   &---  (87.87)  &--- (94.58) & --- (90.30) \\
   \multirow{1}{*}{nnFormer~\cite{zhou2021nnformer}} 
   &90.94 & 89.58 &95.65 &92.06 \\
    \multirow{1}{*}{UNETR++~\cite{shaker2022unetr++}}    
   &\underline{91.89} & \underline{90.61} &\underline{96.00} &\underline{92.83} \\
   \hline \hline
   \multirow{1}{*}{MAE + UNETR++ (ours)}    
   &\textbf{92.59} &\textbf{91.38}&\textbf{96.37}   &\textbf{93.45}  ($0.62\% \uparrow$) \\\hline
   \multirow{1}{*} {$p$-value}
                     & \multicolumn{4}{c}{$< 1e-2$ (Dice)}   \\\hline 
   \end{tabular}
   }
\end{table}

\begin{table}[!t]
   \caption{Segmentation results on the MSD spleen dataset.}\label{tab:spleen}
   \vspace{-4mm}
    \centering
    \vspace{0.2cm}
    \scalebox{0.85}{
   \begin{tabular}{l|cc}
   \hline 
   \multirow{1}{*}{Method}  & \multirow{1}{*}{Dice ($\uparrow$)} & \multirow{1}{*}{HD95 ($\downarrow$)} \\\hline\hline
    \multirow{1}{*}{SETR MLA~\cite{zheng2021rethinking}}    
                     &0.950   & 4.091           \\
\multirow{1}{*}{TransUNet~\cite{chen2021transunet}}   
                     &0.950   & 4.031        \\
     \multirow{1}{*}{AttUNet~\cite{oktay2018attention}}    
                     &0.951   & 4.091          \\
   \multirow{1}{*}{U-Net~\cite{ronneberger2015u}}    
                     &0.953   & 4.087            \\
         \multirow{1}{*}{CoTr~\cite{xie2021cotr}}  
                     &0.954   & 3.860         \\
            \multirow{1}{*}{UNETR~\cite{hatamizadeh2022unetr}}   
                     &{0.964}   & {1.333}           \\
            \multirow{1}{*}{UNETR + MAE~\cite{zhou2022self}}
            &--- (\underline{0.966})   &---  (1.295)  \\
            \multirow{1}{*}{UNETR++~\cite{shaker2022unetr++}}
                     &\underline{0.966}   &\underline{1.246}  \\
   \hline \hline
   \multirow{1}{*}{MAE + UNETR++ (ours)}    
   &\textbf{0.974 $(0.8\% \uparrow)$} &\textbf{1.002 $(0.244 \uparrow)$}    \\\hline
   \multirow{1}{*} {$p$-values}
                     &   \multicolumn{2}{c}{$< 1e-2$ (Dice), $< 5e-2$ (HD95)}  \\\hline 
   \end{tabular}
   }
\end{table}

\begin{table}[!t]
   \caption{Segmentation results on the MSD lung dataset.}\label{tab:lung}
   \vspace{-4mm}
    \centering
    \vspace{0.2cm}
     \scalebox{0.88}{
   \begin{tabular}{l|c}
   \hline 
   \multirow{1}{*}{Method}  & \multirow{1}{*}{Dice ($\uparrow$)} \\\hline\hline
   \multirow{1}{*}{UNETR~\cite{hatamizadeh2022unetr}}   
                      & 73.29           \\
            \multirow{1}{*}{nnU-Net~\cite{isensee2021nnu}}
              & 74.31    \\
            \multirow{1}{*}{Swin UNETR~\cite{hatamizadeh2022swin}}
               & 75.55    \\
            \multirow{1}{*}{nnFormer~\cite{zhou2021nnformer}}
                       & 77.95  \\
             \multirow{1}{*}{UNETR + MAE~\cite{zhou2022self}}
                      & --- (78.90)    \\
            \multirow{1}{*}{UNETR++~\cite{shaker2022unetr++}}
                       &\underline{80.68}\\
   \hline \hline
   \multirow{1}{*}{MAE + UNETR++ (ours)}    
   & \textbf{82.55} $(1.87\% \uparrow)$     \\\hline
   \multirow{1}{*} {$p$-value}
                     & \multicolumn{1}{c}{$< 5e-2$ (Dice)}   \\\hline 
   \end{tabular}
   }
\end{table}

\begin{table*}[t]
    \centering
    \caption{Ablation study of the effects of different key components in our MAE + UNETR++ approach on the Synapse CT dataset.}
    \label{tab:ablation}
    \scalebox{0.72}{
    \begin{tabular}{c|c|c|c|c|c|c}
        \hline
        Method & Pre-trained ViT Encoder & Spatial Pre-text Task & Topological Loss & Pre-trained UNETR++ & Dice ($\uparrow$) & HD95 ($\downarrow$) \\
        \hline
        \makecell{UNETR++}  &  &  & & & 87.22 & 7.53 \\\hline
        \makecell{UNETR++ \\ with pre-trained ViT encoder} & $\surd$ & & & & 87.62 & 7.03 \\\hline
        \makecell{UNETR++ \\ with pre-trained ViT encoder \\ \& spatial pre-text task} & $\surd$ & $\surd$ & & & 88.09 & 6.12 \\\hline
        \makecell{UNETR++ \\ with pre-trained ViT encoder \\ \& spatial pre-text task \\ \& topological loss} & $\surd$ & $\surd$ & $\surd$ & & 88.34 & 5.96 \\\hline
        \makecell{MAE + UNETR++} & $\surd$ & $\surd$ & $\surd$ & $\surd$ & \textbf{88.94} & \textbf{5.89} \\\hline
    \end{tabular}
    }
\end{table*}





\subsection{Experimental Results}\label{exp-results}
\begin{figure}[h!]
    \centering
    \includegraphics[width=0.9\linewidth]{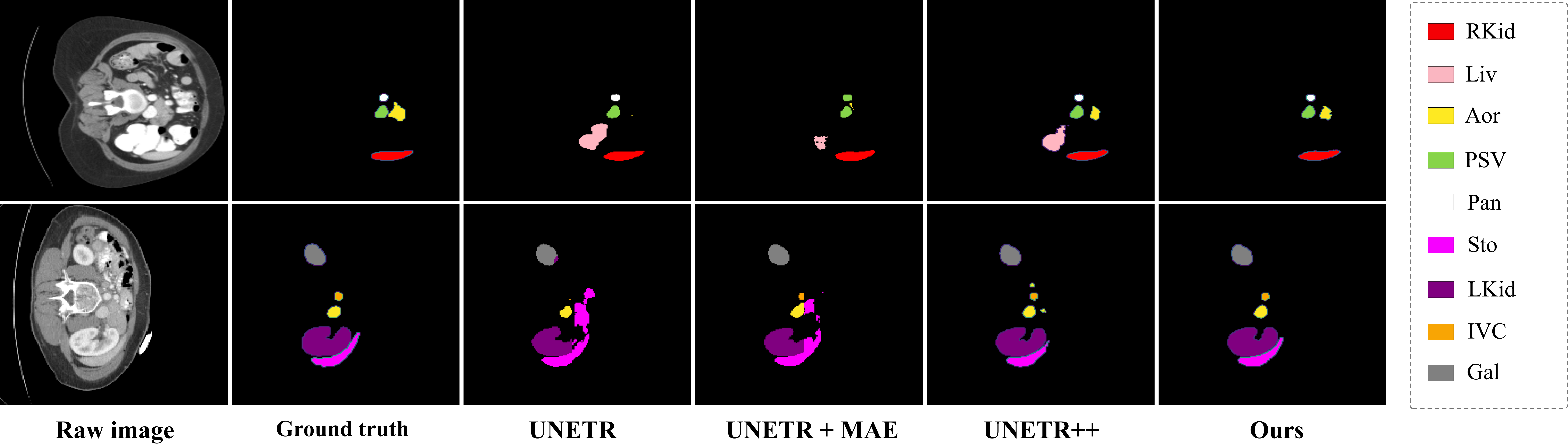}
    \caption{Visual results of different methods on the Synapse CT dataset.}
    \label{vis:sypnase}
\end{figure}

\textbf{Synapse CT Dataset Results.}
In Table~\ref{tab:Synapse}, we compare our method with an array of baseline methods (U-Net~\cite{ronneberger2015u}, TransUNet~\cite{chen2021transunet}, UNETR~\cite{hatamizadeh2022unetr}, Swin-UNet~\cite{cao2023swin}, MISSFormer~\cite{huang2021missformer}, Swin UNETR~\cite{hatamizadeh2022swin}, and nnFormer~\cite{zhou2021nnformer}) and SOTA models (UNETR++~\cite{shaker2022unetr++}, and the MAE-based self pre-training method, i.e., UNETR + MAE~\cite{zhou2022self}). 
On this dataset, UNETR++ yields superior performance over the other known methods. Our method 
outperforms UNETR++ by 1.72\% and 1.64~mm in average Dice and HD95, respectively, which are quite impressive improvements on the Synapse CT dataset.
Specifically, our method achieves the highest Dice scores on six organs (kidney (right), kidney (left), gallbladder, liver, stomach, and aorta).
Compared to the known methods, our method is more advantageous in segmenting gallbladder, which is difficult to delineate using known segmentation methods.
Our method is able to surpass the UNETR + MAE by large margins in both the evaluation metrics, demonstrating the effectiveness of our method (see Fig.~\ref{vis:sypnase}).

\textbf{BTCV CT Dataset Results.}
Table~\ref{tab:btcv} showcases the segmentation results of various methods on the BTCV CT dataset.
Among the known methods, nnU-Net~\cite{isensee2021nnu} and UNETR++~\cite{shaker2022unetr++} achieve average Dice scores of 83.16\% and 83.28\%, respectively.
Our method outperforms the SOTA method UNETR++ by 0.8\% in average Dice. 
This is particularly commendable given the challenging nature of the BTCV CT dataset, which encompasses 13 distinct organs.

\textbf{ACDC Dataset Results.}
Table~\ref{tab:acdc} reports the quantitative results on the ACDC dataset.
We observe that nnFormer~\cite{zhou2021nnformer} and UNETR++~\cite{shaker2022unetr++} attain better performances of 92.06\% and 92.83\% in average Dice, respectively.
Remarkably, our method surpasses the SOTA method UNETR++ by 0.62\% in the average Dice score. 
Furthermore, our method outperforms 
the MAE-based self pre-training method UNETR + MAE~\cite{zhou2022self} by an impressive 3.15\% in average Dice, confirming the effectiveness of our new approach.

\textbf{MSD Spleen Dataset Results.}
As Table~\ref{tab:spleen} shows, on the MSD spleen dataset, both UNETR + MAE~\cite{zhou2022self} and UNETR++\cite{shaker2022unetr++} already achieve a very high 0.966 Dice score, giving a limited margin for big improvement. 
Nonetheless, our method still manages to enhance both the Dice and HD95 scores by 0.8\% and 0.244, respectively. These results reinforce the superiority of our method over known SOTA methods.

\textbf{MSD Lung Dataset Results.}
Table~\ref{tab:lung} presents the experimental results on the MSD lung dataset. 
One can see that the best known method is UNETR++~\cite{shaker2022unetr++}, whose Dice score is higher than the second-best method, UNETR + MAE~\cite{zhou2022self}, by a margin of 1.78\%.
In comparison, our method outperforms UNETR++ by 1.87\% and 
UNETR + MAE by a notable 3.65\% in Dice score. 
These substantial improvements validate the effectiveness of our method.

\subsection{Ablation Study}\label{abla}
We conduct ablation study using the Synapse CT dataset to examine the effects of different key components in our method.
From the results in Table~\ref{tab:ablation}, we observe the following. (1) When applying the pre-trained ViT encoder on top of UNETR++, the Dice is improved by $0.4\%$ ($p$-value $= 0.027$, $t$-test), showing the effect of our fine-tuned model.
(2) When adding the spatial pre-text task to the pre-training of ViT only, the Dice is further improved by $0.47\%$ ($p$-value $= 0.009$, $t$-test), demonstrating the effect of our proposed pre-text task in extracting spatial information. 
(3) When further adding the topological loss to the pre-training of ViT only, the Dice is further improved by $0.25\%$ ($p$-value $= 0.015$, $t$-test), validating the effect of our proposed topological loss in capturing geometric shape information. 
(4) When transferring the weights of both the ViT encoder and UNETR++ to the fine-tuned model, the Dice is improved by $0.60\%$ ($p$-value $= 0.011$, $t$-test), validating the effects of the MAE pre-training strategy on UNETR++ as well as the spatial and reconstruction consistency losses on co-pretraining ViT and UNETR++.

\section{Conclusions}
\label{sec:conclu}
In this paper, we proposed a novel extension of masked autoencoders (MAEs) for self pre-training (i.e., models pre-trained on the same target dataset) for 3D medical image segmentation.
In particular, we proposed a new topological loss for extracting geometric shape information, introduced a pre-text task to aggregate global spatial information, extended the MAE pre-training strategy to a hybrid SOTA medical image segmentation architecture, and developed a fine-tuned model to further improve the downstream segmentation performance.
Experimental results on five public 3D segmentation datasets demonstrated the effectiveness of our proposed approach.

\section{Acknowledgements}
This research was supported in part by NSF Grants CCF-2523787 and OAC-2104158.

\bibliographystyle{plain}
\bibliography{refs}{}
\end{document}